\documentclass[conference]{template/IEEEtran}
\IEEEoverridecommandlockouts

\usepackage{template/macro}

\usepackage[T1]{fontenc}
\usepackage[utf8]{inputenc}
\usepackage[english]{babel}

\usepackage{times}

\PassOptionsToPackage{hyphens}{url}
\usepackage{url}
\usepackage{hyperref}

\usepackage{graphicx}
\usepackage{caption}
\usepackage{tikz}
\usepackage{pgfplots}
\pgfplotsset{compat=1.17}

\usepackage{array}
\usepackage{booktabs}
\usepackage{microtype}
\usepackage{xcolor}
\usepackage{enumitem}
\setitemize{itemsep=0pt,parsep=0pt,topsep=0pt}
\setenumerate{itemsep=0pt,parsep=0pt,topsep=0pt}
\usepackage[ruled,vlined]{algorithm2e}

\usepackage[sectionbib,authoryear,round]{natbib}
\usepackage[normalem]{ulem}

\begin{document}

\title{[Vision Paper]\\ Learning with Variational Objectives on Measures}
\author{\IEEEauthorblockN{Vivien Cabannes}
\IEEEauthorblockA{\textit{Meta AI, FAIR Labs} \\
New York City, USA \\
vivc@meta.com}
}

\maketitle

\begin{abstract}
The theory of machine learning has predominantly focused on variational objectives expressed on functions.
Yet, the advent of big data has exposed limitations in these formulations: functions struggle to capture uncertainty arising from data veracity issues; and the assumption of data independence often conflicts with data variety.
In this vision paper, we propose a shift towards measures as the basis for variational objectives, highlighting their advantages to quantify uncertainty, to generalize out of training distribution, and to deal with scarce annotations.
We aim to inspire and stimulate further research in adapting conventional machine learning algorithms to optimize these new objectives, without compromising statistical efficiency.

\end{abstract}

\begin{IEEEkeywords}
   Statistical learning, distribution learning, measure spaces.
\end{IEEEkeywords}

\section{Introduction}
Historically, machine learning has focused on learning functions.
Statistical learning theory has provided a relatively clear understanding of the error to expect when utilizing those functions based on the number of samples seen at training time.
However, with the rise of big data, machine learning applications have grown large-scale, tackling more complex problems, and the guarantees offered by this theory have lost their sharpness \citep{AIFear0,AIFear1}.
Several factors come into play.
\begin{itemize}
    \item Variety: utilization environments can differ significantly from training and testing ones, those latter sometimes hiding hard-to-mitigating biases \citep{Bias0,Bias1}.
    \item Veracity: datasets are too large to be fully annotated, enforcing engineers to turn to new learning frameworks such as self-supervised \citep{SSL} or weakly supervised learning \citep{Weak}.
    \item Value: the functional point-of-view is not ideal to describe new machine-learned algorithms, such as large language models \citep{LLM0,LLm1} or diffusion models \citep{Dif0,Dif1}, whose ultimate goal is to generate content resembling the input it was trained on.
\end{itemize}

To address these limitations, this paper proposes to cast statistical learning theory into a theory that encompasses variational objectives on measures.
This approach aims to provide a more effective description of real-world machine learning scenarios. 
It enables the design of robust algorithms that can handle data variety, to cope with missing data annotations, and to model large language models and diffusion models ultimate goals.
While this paper exposes the motivation and rationale behind this new framework, it leaves the practical construction of this framework as an open problem.
We believe that exploring this research direction will lead to advancements in addressing the challenges of modern machine learning applications and pave the way for new algorithmic developments.

The paper is organized in three sections.
The first section lays out an abstract vision on learning with variational objectives on measure.
The second section provides research directions to ensure the statistical soundness of those objectives.
The third section discusses the potential applications of such objectives, motivating our vision with many concrete examples.

\section{The Problem}
This section delves into the foundation of measure-based learning theory.
It recalls classical perspectives on learning theory, centered around functions.
It then offers a counterpart based on measures, before exposing open problems whose solutions would advance the frontier of knowledge in learning theory.

\subsection{Classical Learning Theory}
The theory of statistical learning has focused on the learning of functions \citep{Stat0,Stat1,Stat2,Stat3,Stat4}.
In essence, it models any algorithm as a rule to map inputs (in a space $\X$) to outputs (in a space $\Y$).
Learning an algorithm consists in learning a mapping $f:\X\to\Y$ that captures the relationship between inputs and outputs.
It is useful to model data point $(X, Y)$ as coming from a distribution $\rho\in\prob{\X\times\Y}$ over input/output pairs.
This distribution can equally be understood through its marginal distribution over inputs $X\sim\rho_\X \in \prob\X$, which will be fed to the machine at inference time, and the conditional distributions $\paren{Y\midvert X}\sim\rho \vert_x\in\prob\Y$ that generate the outputs $y$ given the inputs $x$, and that we would like the machine to be able to predict.
The notion of optimal mapping is defined through the variational objective
\begin{equation}
\label{eq:risk}
    f^* \in \argmin_{f:\X\to\Y} \cR(f):=\E_{(X, Y) \sim \rho}[\ell(f(X), Y)],
\end{equation}
where $\rho \in \prob{\X\times\Y}$ is the resulting distribution on input-output pairs,\footnote{%
    To build $\rho$ from the marginal $\rho_\X$ and the conditionals $(\rho\vert_x)_{x\in\X}$, recall that $\E_{(X,Y)\sim\rho}[F(X, Y)] = \E_{X\sim\rho_\X}\bracket{\E_Y\bracket{F(X, Y)\midvert X}}$, and that we also have  $\E_Y\bracket{F(x, Y)\midvert X=x} = \E_{Y\sim\rho\vert_x}[F(x, Y)]$.
} and $\ell:\Y\times\Y\to\R$ quantifies the error cost $\ell(f(x), y)$ associated with predicting $f(x)$ when the desired output was actually $y$.

In practice, to minimize the risk $\cR$, it is usual to collect $n$ examples $(x_i, y_i)$ that are thought of as independent realizations of the random variable $(X,Y)\sim\rho$, and to minimize the empirical approximation
\begin{equation}
    f_n \in \argmin_{f\in\cF} \cR_n(f):= \frac{1}{n}\sum_{i=1}^n[\ell(f(x_i), y_i)],
\end{equation}
where $f$ is constrained to belong to a class of function $\cF$ which implicitly encodes how the learned function $f$ should generalize to unseen example $X\sim\rho_\X$ based on the training examples $(x_i, y_i)_{i\in[n]}$.

This perspective has laid the foundation of learning theory.
It has established a solid ground to design learning procedures based on training data, particularly in the context of relatively small datasets.
However, as machine learning applications have expanded in the era of big data, new uncharted territories have emerged, requiring new tools and methodologies.

\subsection{Measure-based Statistical Learning}
In this vision paper, we propose revisiting the skeleton of statistical learning in order to apply it to measure-based variational objectives.
Instead of learning the mapping $f:\X\to\Y$, we shift our attention to learning a distribution $\mu\in\cZ$ over some space $\cZ$.
This is achieved through a principled population objective as
\begin{equation}
    \mu^* \in \argmin_{\mu\in\prob\cZ} \cE(\mu); \qquad \cE:\prob\cZ\to\bar\R,
\end{equation}
where $\bar\R = \R \cup \brace{+\infty}$. 
Hear our goal is to find the optimal distribution $\mu^*$ that minimizes a function $\cE$ over measure.
Note that any hard constraints $\mu \in \cC$ can be added to the objective $\cE$ by adding a barrier regularization $\chi_\cC$, where $\chi_\cC(\mu) = 0$ if $\mu$ satisfy the constraints $\mu \in \cC$ and $\chi_\cC(\mu) = +\infty$ otherwise.

In practical scenarios, it might not be feasible to directly access this population objective $\cE$ and work with its exact form.
Instead, one may only be able to work with an approximation $\hat\cE$ of the objective.
This approximation may arise due to modeling limitations, to the sole access of samples instead of distribution, or to optimization constraints that necessitate the usage of a surrogate objective.
As a consequence in practice, one might only be limited to optimize for
\begin{equation}
    \mu_n \in \argmin_{\mu\in\prob\cZ} \cE_n(\mu); \qquad \cE_n:\prob\cZ\to\bar\R,
\end{equation}
where $\cE_n$ is an approximation of the original $\cE$.

\subsection{Open Problems}
Our vision leads to the following theoretical problem: 
\begin{quote}
{\em Can we define abstract classes of objectives $\cE$ and of their (random) approximations~$\cE_n$, such that $\cE(\mu_n)$ converges to $\cE(\mu^*)$ as $n$ goes to infinity? 
Can we give rates of convergence of such a convergence under some well-specified model assumptions? 
Do those abstract classes describe problems of practical interest?}
\end{quote}
Addressing these questions would give insights into the conditions under which one could design measure-based variational frameworks of practical interests.
This analysis would shed light on the applicability and practical utility of these objective classes and guide the development of innovative methodologies in machine learning.

Those questions could equally be explored from the practical side. 
\begin{quote}
{\em Can we define objectives $\cE$ and of their (random) approximations~$\cE_n$, such that trying to minimize $\cE_n$ leads to distribution $\mu_n$ with favorable practical properties?
Can we find examples where this perspective advances the state-of-the art of a machine learning application?}
\end{quote}
Finding successful examples would provide crucial clues into a generic learning framework to uncover, paving the way to a holistic approach to tackle a wide range of practical problems.
Overall, the discovery of a successful measure-based variational framework would help to chart the landscape of machine learning in the big data era. 

\section{Theoretical Research Directions}

This section proposes directions to lay the theoretical foundations of learning with measure-based approaches.

\subsection{Following the classical path}
A natural direction to establish convergence guarantees of a measure-based approach is to follow the path of classical statistical learning.
First of all, remarking that $\cE_n(\mu^*) \geq \cE_n(\mu_n)$ leads to the inequality
\[
    \cE(\mu_n) - \cE(\mu^*) \leq \cE(\mu_n) - \cE_n(\mu_n) + \cE_n(\mu^*) - \cE(\mu)
\]
When $\cE_n$ is a random variable, the convergence of $\E[\cE_n(\mu)]$ to $\cE(\mu)$ is usually simple to establish through variations of the law of large numbers; the speed of the convergence can be derived with variations of central-limit theorem, or more generally the tail of $\cE_n(\mu) - \cE(\mu)$ can be studied with concentration inequalities.
The part in $\cE(\mu_n) - \E\cE_n(\mu_n)$ is more tricky as $\mu_n$ is also a random variable.
It is classical to bound this term with $\sup_{\mu \in\cC} \E[\cE(\mu_n) - \cE_n(\mu_n)]$ for $\cC\subset \prob\cZ$ a set of measure that $\mu_n$ is known to belong to.
From there, one could cover the space $\cC$ with some balls, and control the deviation of interest on each ball, before performing some union bound, eventually with ``chaining’’ to reduce redundant events in the union bound \citep{Stat5}.

\subsection{Lifting measures into functions}
When the space of probability is endowed with a Hilbertian metric \citep{Class2}, the objective
\(
    \E_Z[\ell(\mu,\delta_Z)]
\)
with $\ell(\mu, \mu') = \norm{\mu - \mu'}^2$ is minimized for $\mu = \E_Z[\delta_Z]$.\footnote{%
    Similar statements are true for other ``smooth losses'' \citep{Class3}.
}
When the law of $Z$ is replaced by an empirical approximation, concentration inequalities can be used to ensure the convergence of the resulting $\mu_n$ towards $\mu^*$ in probability with respect to the underlying Hilbertian metric \citep{Class4}.
In some sense this approach lifts a measure space into a function space where outputs are linear combinations of atoms $\delta_z$.
This is reminiscent of classification problems, which are characterized by $\ell(y, z) = \ind{y\neq z}$ in \eqref{eq:risk} \citep{Class0}, where categorical outputs $y\in\Y = \brace{1, \ldots, m}$ are encoding as a one-hot vector $\phi(y) = (\ind{y=y'})_{y'\in\Y}$ \citep{Class1}, and the minimization of the mean square error with the data $(X, \phi(Y))$ leads to the estimate
\(
    f^*(x) = (\Pbb\paren{Y=y\midvert X=x})_{y\in\Y}.
\)

\subsection{Some interesting facts on sigma-fields}
In essence, the fact that one can not directly get $\mu^*$ but has to satisfy oneself with $\mu_n$ is due to some coarseness of the approximation of $\cE$ by $\cE_n$ which does not allow to retrieve as much information as in the ideal case.
The notion of coarseness of information is naturally captured by sigma-fields.
Let $\Sigma_n$ be a sigma field on $\X$. 
Ideally, for any $f:\X\to\R$, we would like $\E\bracket{f\midvert \Sigma_n}$ to be $f$ itself.
Yet in some cases, $\Sigma_n$ might not be rich enough, and one will only retrieve coarse information on $f$ through $\E\bracket{f\midvert \Sigma_n}$ (for example, this could be due to missing information, or only accessing weak information as detailed in the next section).
As $\Sigma_n$ grows larger, this approximation will get finer.
A natural way to build $\cE_n$ would be by considering $\cE_n = \cE + \chi_{\Sigma_n}$ where $\Sigma_n$ converges to the Borel algebra on $\X$ in some sense \citep[e.g.][]{Kudo0}.
In particular, if $\cE$ displays some continuity properties with respect to this convergence topology \citep[see e.g.][for concrete examples]{Kudo1}, this would ensure that convergence properties on the $\Sigma_n$ can be cast on the difference $\cE(\mu_n) - \cE(\mu)$.

\subsection{Looking at the abstract topological picture}
At an abstract level, one could try to leverage abstract topological facts on measure spaces\footnote{%
    While there exists many measure to quantify the distance between measures \citep{Measure0}, some of them metricizing the weak convergence of measures \citep[see Theorem 6.9 in][for example]{Measure1}, ultimately it is sufficient to focus on the topology induced by the objective function $\cE$.
} and on specific variational objectives in order to directly draw conclusions on the good behavior of some methods that work with variational objectives on measures.
For example, when $\cE_n$ is deterministic, the consistency of the estimator $\mu_n$ is nothing but the epi-convergence of $\cE_n$ towards $\cE$ \citep{Epi0,Epi1}.

Interestingly, such a procedure does not only identify the solution of the pure classification problem, but equally provides estimates on how much an input $x\in\X$ is expected to produce different outputs $y\in\Y$.

\section{Practical Examples}
The goal of our vision paper is not only to build an abstract mathematical framework, but also to make sure that such a framework actually helps advance the state-of-the-art of machine learning by answering the need to formulate better objectives on concrete practical problems.
This section gives examples of the variational formulations of important machine learning problems with a measure-based objective.

\subsection{Generative modeling}
The first type of problems that are natural to express with a variational objective on distributions are sampling problems, where one wants to learn to generate samples according to a given distribution $\mu^*$ while only accessing samples $(z_i) \sim (\mu^*)^{\otimes n}$.
The original samples could be high-resolution natural images to train image generators, or high-quality sentences found on the internet to train LLMs.
For example, one can look for the following population principle
\[
    \mu^* \in \argmin_{\mu \in \prob\cZ} D(\mu^* \| \mu)
\]
which could be translated into the empirical approximation
\[
    \hat\mu \in \argmin_{\mu \in \prob\cZ} D(\frac{1}{n} \sum_{i=1}^n \delta_{z_i} \| \mu) + \chi_\cC(\mu),
\]
where $D:\prob\cZ\times\cZ\to\R_+$ is some divergence, the $(z_i)$ are independent realization sampled accordingly to $\mu^*$, and $\cC \subset \prob\cZ$ capture a class of measure that $\hat\mu$ is searched in.

Currently, approaches to generative AI are distinct when it comes to text or images.
Regarding natural language processing \citep{LLM0,LLm1}, large language models learn to generate a sentence one token after the other with a model that assigns probability scores to the different potential completion tokens.
For images, it consists in learning a function $f$, that maps a random variable $Z$ that one knows how to generate (e.g., $Z$ is Gaussian) into a random variable $X$ whose law resembles the target measure $\mu$.
In GANs \citep{GAN}, $f$ is learned with some discriminator to make sure that the generated sample $f(Z)$ are indistinguishable from original ones $X\sim\mu$.
In flow matching \citep{FLOW}, $f$ is learned by integrating a vector field that has been fitted to guarantee the consistency of this approach, replicating the solution of an ordinary differential equation.
Diffusion models \citep{DDPM} provide a slight variation of this process: $f$ is split into different modules, with noise added to each modules in order to simulate the evolution of some reserve stochastic process, replicating the solution of an stochastic differential equation.
Those approaches might provide useful clues to develop a generic framework concerned with learning distributions.
On the other hand, clear statistical guidelines to generalize from empirical samples to population distribution would prove useful to counter memorization problems, where $f$ only generates samples that were already seen at training time.


\subsection{Uncertainty Quantification}
A second natural reason to look at the prediction of distributions instead of functions is given by the practical approach to classification, where rather than learning a function from $\X$ to a set $\cY$ of classes, practitioners prefer to learn a ``score'' function from $\X$ to $\prob\cY$ \citep{Score0,Score1,Score2}.
Such a procedure does not only identify the solution of the pure classification problem (defined with the 0-1 loss $\ell(y, z) = \ind{y\neq z}$), but equally provides estimates on how much an input $x\in\X$ is expected to produce different outputs $y\in\Y$.
Indeed, even in regression settings, it might be preferable to learn $\paren{Y\midvert X}$, the law of $Y$ conditioned on $X$, instead of only learning the conditional expectation $\E\bracket{Y\midvert X}$, since this could help to quantify the uncertainty that the output associated with $X$ should be $f(X)$.
As such, formulating learning at the level of measure offers a more comprehensive approach to extract meaningful insights and make informed predictions regarding big data filled with noisy information.
Recent advances in conformal prediction have provided interesting directions to leverage when formulating the problem of quantifying a predictor uncertainty from a measure point of view \citep{IC0,IC1}.

\subsection{Out-Of-Distribution Generalization}
While machine learning usually needs curated data in order to be trained efficiently, one may be interested in using the learned algorithm under a distribution which is quite different from the training one.
This is particularly true in the era of big data, where the likelihood of encountering shifts in distributions increases significantly.
There are two potential sources of variations between training to utilization settings: either the conditional distributions $\rho\vert_x$ are changing, which might be due to a change of confounding variables (e.g., $x$ corresponds to geographic locations, $y$ to some atmospheric conditions, but training took place in winter only), or the marginal $\rho_\X$ is changing (e.g., training took place in the Western world, but utilization is done worldwide), which is known as covariate shift \citep{Shift}.
Eventually, one could have a prior $\mu \in \cC$ on future distributions in utilization settings, and try to minimize the worse performance over utilization environment,
\begin{align*}
    \mu &\in \argmin_{\mu\in\prob\cZ} - \inf_{f:\X\to\Y} \E_{(X,Y)\sim\rho}\bracket{\ell(f(X), Y)} + \chi_\cC(\mu)
    \\&= \argmax_{\mu\in\cC} \cR^*(\mu),
\end{align*}
where $\cR^*(\mu)$ is the minimum risk in \eqref{eq:risk} one can suffer when the data are sampled accordingly to $\mu$.

\subsection{Weak Information Disambiguation}
A final example is provided by a situation where one only has access to weak information on the real data distribution $\rho$.
This could be due to missing data \citep{Missing}, or to only accessing label proportions \citep{Prop}, weak supervision on the labels \citep{Weak}.
Once again, this problem can be formulated as needing to find a candidate measure for what $\rho$ should be, given only underdetermined constraints $\rho \in \cC \subset \prob\cZ$ provided by the weak information.
In this setting, it is natural to fill the missing information by looking at the simplest distribution in the set of potential distributions.
This notion of simplicity can eventually be captured by the entropy of the distribution, leading to 
\[
    \mu^* \in \argmin_{\mu \in \cC} H(\mu) := \E_{Z\sim\mu}[\log\paren{\frac{\diff \mu}{\diff \tau} (Z)}],
\]
for some base measure $\tau\in\prob\cZ$, with $\diff \mu / \diff \tau$ denoting the Radon–Nikodym derivative.
Interestingly, this has been the approach taken by \citet{Partial}, who capture the notion of simplicity through a loss-based variance defined as
\[
    \cE(\rho) = \inf_{f:\X\to\Y} \E_{(X,Y)\sim\rho}[\ell(f(X), Y)],
\]
and leverage kernel mean embedding techniques to translate this objective into an empirical one.
The need to better understand learning with missing data or incomplete information is crucial for big data, especially when data is collected from various heterogeneous sources.

\section{Conclusion}
In this vision paper, we have explained how it could be beneficial to derive a new theory of statistical learning, where instead of focusing on the input-output mappings, one focuses on distributions.
In particular, it would be convenient to derive a mathematical framework to cast principled population objectives on measures into approximate objectives (the approximation being eventually due to limited access to empirical data samples or to computation constraints), while making sure that as the approximation gets finer, one retrieves an optimal measure.
We have motivated such a theory with several examples where it could have a deep impact to solve some of the key challenges of nowadays machine learning, which has entered the era of big data.

\section*{Acknowledgment}
V. C. would like to thank L\'eon Bottou for raising a question on out-of-distribution generalization, and mentioning Vladimir Vapnik's interest to foster his theory with a measure oriented point-of-view.

\bibliography{main}

\end{document}